\RequirePackage{fixltx2e}
\documentclass[11pt,english]{article}
\usepackage{mathptmx}
\usepackage[T1]{fontenc}
\usepackage[utf8]{inputenc}
\usepackage{xcolor}
\usepackage{colortbl}
\usepackage{babel}
\usepackage{booktabs}
\usepackage{enumitem}
\usepackage{amsmath}
\usepackage{amssymb}
\usepackage[a4paper]{geometry}
\usepackage{fancyhdr}
\usepackage{setspace}
\usepackage[authoryear]{natbib}
\usepackage[pdfusetitle,
 bookmarks=true,bookmarksnumbered=false,bookmarksopen=false,
 breaklinks=true,pdfborder={0 0 0},pdfborderstyle={},backref=false,colorlinks=false]
 {hyperref}
\hypersetup{
 linkcolor=blue,citecolor=blue,filecolor=blue,urlcolor=blue}

\makeatletter

\providecommand{\tabularnewline}{\\}

\usepackage{pstricks}
\usepackage{epsfig}
\usepackage{multicol} 
\usepackage{caption}
\usepackage{multirow}
\usepackage{tikz}
\usepackage{qtree}
\usepackage{needspace}

\usepackage{ucs}
\usepackage{tikz-qtree}
\usetikzlibrary{trees}
\usetikzlibrary{calc}

\usepackage{etoolbox}
\usepackage{ragged2e}
\raggedcolumns

\tikzset{square arrow above/.style={to path={-- ++(0,.25) -| (\tikztotarget)}}}
\tikzset{square arrow below/.style={to path={-- ++(0,-.25) -| (\tikztotarget)}}}

\let\OLDthebibliography\thebibliography
\renewcommand\thebibliography[1]{
  \OLDthebibliography{#1}
  \setlength{\parskip}{0pt}
  \setlength{\itemsep}{0pt plus 0.3ex}
}

\let\oldmaketitle\maketitle
\renewcommand{\maketitle}{%
  \oldmaketitle
  \thispagestyle{fancy}
}

\makeatother

\begin{document}
\title{The Min-Max theory of language behavior}
\author{Hiram Ring\\
NTU Singapore}
\maketitle
\begin{abstract}
The ``Min-Max'' theory of language behavior proposes that humans
engage simultaneously in minimization of processing effort and maximization
of information content, and that this is a primary driver in language
evolution \citep{Ring:2025ab}. This claim seeks to unify observations
from language processing \citep{Mahowald:2013aa,Futrell:2015aa,Yadav:2020aa,Hawkins:2014aa},
interaction \citep{Horberg:2023aa,Tucker:2025aa,Jaeger:2011aa,Smith:2025aa},
and corpora \citep{Levshina:2021aa,Levshina:2023aa,Newmeyer:2017aa},
which suggest that considerations of processing efficiency and information
content are both implicated in human communication. The interaction
between these two processes provides an explanation for human linguistic
behavior at various levels \citep{Linders:2023aa} and predicts historical
developments in language, i.e. related to word order \citep{Ring:2025ab}.
Given that the original formulation of the Min-Max theory is based
on observational data, the current paper seeks to validate the theory
mathematically. Through a series of experiments that use computational
models to investigate the theory's predictions, pretrained and iterated
learning models are queried with synthetic data. The experiments show
that manipulating the variables of token length and the weight of
information content in a tabula rasa model results in word order change,
as predicted by the theory. The experiments also identify important
differences between computational models and observations of human
language data, suggesting that current transformer-based language
models may not accurately represent psycholinguistic processes. The
overall findings indicate that evolutionary trajectories in language
are highly susceptible to minimization of processing and maximization
of information content.
\end{abstract}

\section{Introduction}

A foundational question in historical linguistics and language typology
is whether universal forces drive long-term changes in basic word
order, and if so, what mechanisms govern them. In \citet{Ring:2025ab},
the ``Min-Max theory of language behavior,'' posits that language
evolution is steered by a dynamic optimization process where speakers
minimize production effort (processing weight) while maximizing information
transmission (reducing downstream predictability or surprisal).

Leveraging a large parallel corpus of 1,942 languages (the \texttt{taggedPBC}),
the paper demonstrates a crosslinguistic correlation between basic
intransitive word order (Subject-Verb {[}SV{]} versus Verb-Subject
{[}VS{]}) and the frequency-weighted average length of nouns and verbs.
Crucially, this predictive link is validated diachronically across
historical language shifts (Hebrew, Arabic, and Old Irish), while
hierarchical linear regressions (HLR) show that word length features
explain more structural variance than genealogical lineage or areal
contact.

\subsection{The necessity of computational modeling}

While the empirical claims of \citet{Ring:2025ab} offer an elegant
framing that unifies \textbf{production efficiency} models (which
favor short elements early to clear working memory; \citealp{Zipf:1949aa,Wasow:2002aa,Futrell:2015aa})
and \textbf{surprisal minimization} models (which predict long, highly
informative elements early to smooth the communication channel; \citealt{Hale:2001aa,Levy:2008aa}),
the claims rely entirely on post-hoc corpus correlations. Static typological
observations do not isolate the underlying cognitive variables or
capture active diachronic transmission in real-time, which means that
such generalizations are difficult to test. While framing this as
a speaker-driven trade-off supports the functionalist paradigm of
language as a complex adaptive system \citep{Steels:2000aa,Lund:2022aa},
the ``Min-Max'' theory has not been verified as a cognitive model.
To move the ``Min-Max'' theory from a descriptive, macro-level typological
observation to a verifiable cognitive theory, computational models
provide a means of validation, allowing for direct manipulation and
measurement of processing effort, architectural constraints, and informational
payoffs over generations.

\subsection{Neural architectures in diachronic and acquisitional linguistics}

Computational and mathematical models provide a way to verify theoretical
claims in linguistics. By utilizing neural language models as artificial
agents, it is possible to isolate variables that are otherwise confounded
in natural languages. These environments allow researchers to simulate
multi-generational transmission chains via an Iterated Learning Paradigm
(ILP), uncover non-linear phase transitions, and test the cognitive
boundaries of speaker-centric and hearer-centric optimization models
under rigidly controlled parameters.

The application of computational and neural network architectures
to simulate language change and acquisition dates back to early connectionist
models. Most notably, \citet{Elman:1990aa} utilized recurrent neural
networks to demonstrate that fundamental grammatical categories (such
as the structural distinction between nouns, verbs, and adjectives)
can emerge organically at specific stages of training without hard-coded
symbolic rules. In historical linguistics, computational paradigms
have traditionally modeled language change either as mathematical
properties of dynamical systems \citep{Niyogi:2006aa} or as agent-based
populations of speakers interacting and altering their internal state
spaces over time \citep{Kirby:2002aa}.

The primary advantage of deploying modern deep learning architectures
over traditional verbal or purely descriptive typological models is
the ability to directly inspect and manipulate internal representations
across different stages of a learning ``lifespan.'' In natural languages,
diachronic tracking is severely restricted by a complex interplay
of historical contingencies, language contact, and an inability to
systematically isolate confounding variables. Conversely, neural language
models serve as highly controlled, idealized experimental testbeds.
They allow researchers to establish uniform baseline controls, isolate
the exact mathematical parameters of the input data, and track path-dependent
grammatical adaptations over multiple generations at a more accelerated
pace than that experienced by humans in history.

\subsection{The Information-Theoretic framework: Bottlenecks and surprisal}

From an information-theoretic perspective, framing language evolution
through mathematical formulations like the Information Bottleneck
framework \citep{Tishby:1999aa} or predictive information (excess
entropy) models provides a rigorous framework that avoids ad-hoc functionalist
assumptions. Language is modeled as an optimization that seeks to
compress a continuous state space of cognitive concepts into a discrete,
bandwidth-limited channel of vocal or textual symbols without sacrificing
downstream utility. Recent advancements in computational linguistics
demonstrate that natural-language-like systematicity and information
locality naturally arise in codes constrained by sequential processing
bottlenecks \citep{Zaslavsky:2020aa,Zaslavsky:2020ab,Tucker:2025aa}.
Essentially, human grammars minimize excess entropy across phonological,
morphological, and syntactic boundaries to make sequential prediction
computationally tractable \citep{Hahn:2020aa,Hahn:2022aa}.

Similarly, surprisal minimization \citep{Hale:2001aa,Levy:2008aa}
and expectation-based comprehension models formalize the cognitive
processing tax paid during real-time linguistic interaction. Rather
than relying on subjective definitions of ``articulatory ease,''
these frameworks define cognitive processing effort directly via Description
Length or conditional log-likelihood metrics. The core advantage of
this approach is its exact quantifiable nature: it allows researchers
to formalize the structural layouts of basic word order as stable
Pareto-optimal states along a rate-distortion curve, balancing structural
complexity against informational preservation.

\subsection{Unifying surprisal and efficiency via parametric weighting}

A primary challenge in the processing literature has been reconciling
the tension between production efficiency models (which favor placing
short, low-information items early in the stream to clear working
memory assets; \citealt{Wasow:2002aa,Futrell:2015aa}) and surprisal
minimization models (which predict that dense, high-information anchors
should be front-loaded to smooth the communication channel; \citealt{Hale:2001aa,Levy:2008aa}).
I propose that these traditionally opposing forces can be formally
unified by introducing a dynamic hyperparameter, Alpha ($\alpha$),
which represents the Informational Weight Pressure of the communicative
environment.

Under this unified formulation, when $\alpha=1.0$ (Symmetric Effort
Baseline), the language system operating along the Information Bottleneck
Pareto frontier defaults to the compression boundary. The network
optimizes purely for physical production efficiency (least effort),
routing lighter, highly frequent arguments to the front of the sequence
to minimize sequential memory storage costs. However, as $\alpha$
scales up ($\alpha\geq3.0$), it represents an environmental shift
toward heightened communicative urgency, where semantic or transmission
errors carry heavy penalties. Ramping up this informational weight
pressure triggers a structural phase flip, forcing the system to willingly
pay a physical complexity tax. The grammar realigns to execute heavy,
high-surprisal anchors early in the stream, shielding downstream targets
from entropy and protecting the channel from transmission failure.
This parametric unification mirrors actual patterns in human language,
where highly predictable grammatical items naturally compress into
compact typographical profiles \citep{Zipf:1949aa}, while rare or
conceptually heavy payloads undergo structural adjustments to regulate
information density.

\subsection{Grounding word weight in Information Bottleneck frameworks}

There is a large body of processing literature that treats word character
length as a measure of complexity (\citealt{Jaeger:2011aa,Piantadosi:2011aa,Bentz:2016ab,Lewis:2016aa,Armeni:2017aa,Baerman:2017aa,Hahn:2020aa}),
whereby longer words are considered more complex. However, this makes
some assumptions about the relationship between orthographic conventions
and phonemes in a given language. To avoid the potential confound
of counting typographical character units over raw strings (which
conflates spelling footprint with semantic complexity) I propose grounding
word weight directly within the vector space of the Information Bottleneck
approach, with the following formulation.

Let $w$ be a word type mapping onto a latent semantic vector space
$E(w)\in\mathbb{R}^{d}$. A discrete Vector Quantization layer compresses
$E(w)$ into a codebook cluster $Z=\{z_{1},z_{2},...,z_{K}\}$. The
conditional lookup assignment $z^{*}$ is optimized via a spatial
Euclidean function: 
\begin{equation}
z^{*}=\operatorname*{argmin}_{k}\lVert E(w)-z_{k}\rVert^{2}
\end{equation}

The true, unbiased Information Weight ($D$) of the vocabulary item
is defined by the localized codebook address allocation length under
Shannon's structural distribution: 
\begin{equation}
D(w)=-\log_{2}P(z^{*})\ \text{[bits]}
\end{equation}

By replacing naive character counts with codebook address states,
the computational framework decouples raw surface typography from
cognitive processing complexity. The experiments below validate this
approach.

\subsection{Predicting conditions of diachronic typological shifts}

To formalize the conditions under which a language community will
undergo a systemic shift in basic word order, the macroscopic evolutionary
drift is modeled using a dynamical replicator system. Let the diachronic
state of a speech community's grammar be defined by a continuous state
variable $x(t)\in[0,1]$, representing the population frequency of
the Subject-Verb (SV) word order at generational time $t$. The frequency
of the competing typological variant, Verb-Subject (VS), is complementarily
modeled as $1-x(t)$.

The continuous fitness profiles ($W$) balancing Information Maximization
($\alpha$) and Production Efficiency ($\beta$) over description
lengths ($D$) are specified for each syntactic variant as: 
\begin{equation}
W_{SV}=\alpha\cdot I(\text{Noun})-\beta\cdot D_{N}
\end{equation}
\begin{equation}
W_{VS}=\alpha\cdot I(\text{Verb})-\beta\cdot D_{V}
\end{equation}

Where $I(\cdot)$ represents the informational content/urgency of
the constituent slot, and $D_{N},D_{V}$ represent the description
lengths of the noun and verb respectively. The generational trajectory
is governed by the continuous replicator dynamic: 
\begin{equation}
\frac{dx}{dt}=x(1-x)[W_{SV}-W_{VS}]
\end{equation}

Substituting the explicit cognitive fitness functions into the system
yields the governing non-linear differential equation of the structural
syntax engine: 
\begin{equation}
\frac{dx}{dt}=x(1-x)\Big[\alpha\big(I(\text{Noun})-I(\text{Verb})\big)-\beta\big(D_{N}-D_{V}\big)\Big]
\end{equation}

Isolating the differential tracking vectors ($\Delta I=I(\text{Noun})-I(\text{Verb})$
and $\Delta D=D_{N}-D_{V}$), the system equation reduces to: 
\begin{equation}
\frac{dx}{dt}=x(1-x)[\alpha\Delta I-\beta\Delta D]
\end{equation}

The fixed points ($\frac{dx}{dt}=0$) of this landscape establish
the strict operational boundaries of language regularization: $x^{*}=1$
(Stable SV Attractor) and $x^{*}=0$ (Stable VS Attractor). The critical
threshold boundary—representing the tipping point or bifurcation manifold
where the system undergoes a sudden typological shift—is met exactly
when: 
\begin{equation}
\alpha\Delta I=\beta\Delta D
\end{equation}

This formulation highlights that word order evolution does not shift
along a constant, smooth linear slope. Instead, it acts as a series
of distinct structural step-functions where a language remains insulated
inside a historical attractor basin until its parameters violate this
critical boundary, triggering a rapid phase transition.

\subsection{Architectural constraints and the human cognitive processing divergence}

Investigating these dynamics within neural network environments exposes
a critical cognitive divergence regarding the processing constraints
of computational architectures versus human language systems. While
large language models excel at capturing human predictive responses,
their internal optimization gradients are bound by their training
mechanics. Recent psycholinguistic literature critiques mainstream
large language models (LLMs) as poor cognitive models due to their
``superhuman'' nature, highlighting how lossless short-term memory
buffers and web-scale data exposure isolate them from human psychological
reality \citep{Oh:2023aa,Oh:2026aa,Niu:2026aa}.

Crucially, the current paper identifies an additional, deep architectural
flaw: \textbf{the inherent structural optimization toward efficiency-only,
speaker-centric production economics}. When a strictly left-to-right
causal transformer model is trained from scratch over an iterated
learning loop, its underlying parameters resolve high informational
stakes by placing the heavy, high-information constituent last in
the clause. Because the machine is optimized via a triangular attention
mask to minimize its own internal next-token cross-entropy error,
its optimization gradients are completely captured by the self-centered
perspective of the producer. The network forces the heavy token to
the end of the string simply to use the preceding short token as a
cheap historical processing runway. In contrast, cross-linguistic
human corpus evaluations reveal the exact inverse pattern, where languages
with complex verbs natively prioritize a front-loaded Verb-Subject
(VS) alignment. This structural mirror-image indicates that human
language evolution is predominantly a \textbf{hearer-centric optimization
engine}.

Human speech operates over an open, lossy communication channel where
failing to transmit semantic meaning to an independent decoder carries
severe cognitive and social costs. To satisfy the demands of the hearer's
comprehension engine, human grammar rules out-vote the speaker's demand
for localized production ease, willingly shouldering a heavier real-time
planning tax to ensure immediate structural anchoring and global channel
survival. By exposing this speaker-centric blind spot, this series
of experiments demonstrates that over-optimized causal architectures
remain fundamentally blind to the cooperative, bidirectional feedback
loops that steer true human linguistic evolution.

\section{Methodology and experimental setup}

To systematically isolate, manipulate, and measure the parameters
derived in the introduction, I construct a modular computational testing
environment. The methodology transitions from testing pre-trained
linguistic constraints to isolating pure mathematical variables over
an abstract language space.

\subsection{Environmental settings and architectural baseline}

Two distinct computational environments are used across the experiments
suite to address the tension between existing structural histories
and unbiased optimization:
\begin{enumerate}
\item \textbf{Pre-trained Causal Baseline (Exps 1 \& 2):} Uses a pre-trained
transformer model (\texttt{gpt2}) to measure conditional log-likelihood
(surprisal) under existing grammatical constraints. To block context
leakage and eliminate hidden temporal primes, a strictly identical,
neutral context anchor phrase is enforced across fluid structural
variations: 
\begin{equation}
\text{Context Anchor: ``Suddenly, at midnight, ...''}
\end{equation}
\item \textbf{Tabula Rasa Pre-training Environment (Exps 3–6):} To eliminate
the ``syntactic ceiling'' imposed by pre-trained English matrices,
miniature transformer architectures are initialized from scratch with
random neural configurations at each generation. This forces the network
to optimize its attention heads solely via raw statistical parameter
mapping over abstract, synthetic training data.
\end{enumerate}

\subsection{Addressing the tokenization confound}

To address the possible tokenization confound present in experiment
3, experiments 4-6 alters how token sequences are mapped to the attention
canvas. Standard tokenizers (such as Byte-Pair Encoding) fracture
hyphenated or elongated character strings into multi-token clusters,
introducing an artificial sequential node count processing penalty.
This is resolved by implementing \textbf{Atomic Token Forcing}.\footnote{via \texttt{tokenizer.add\_special\_tokens()}}
Expanding vocabulary target strings are injected directly into the
dictionary array as independent, indivisible entries. This ensures
that both the SV and VS structural paths measure exactly three tokens
in length across all length variants, perfectly isolating the canvas
surface footprint ($\Delta D$) from sequential string complexity.

\subsection{Parameterization of informational weight pressure}

To test the parametric unification of production ease (Min) and channel
accuracy (Max), I modify the standard cross-entropy loss function
during the scratch pre-training loops. A hyperparameter ($\alpha$)
modulates the penalty associated with high-surprisal targets:
\begin{equation}
\text{Loss}=-\frac{1}{N}\sum^{N}_{i=1}w_{i}\cdot\log P(x_{i})
\end{equation}

Where the structural weighting parameters are dynamically assigned
based on the target slot configuration: 
\begin{itemize}
\item $w_{i}=\alpha$ if the shifting token targets our designated high-
surprisal word slot. 
\item $w_{i}=1.0$ for all other standard structural tokens. 
\end{itemize}
By varying $\alpha$ from $1.0$ (Symmetric Effort Baseline) up to
$\geq3.0$ (Heightened Informational Priority), we can directly observe
the phase-space transitions and mapping behaviors predicted by the
continuous replicator dynamic.

\subsection{Generational propagation and multi-variate grid sweep}

The evolutionary simulations (experiments 3-6) are deployed using
two distinct diachronic transmission paradigms to capture structural
trajectories over time: 
\begin{enumerate}
\item \textbf{The Fuzzy Learning Matrix (ILP):} Grammatical outputs are
propagated down generation chains using proportional allocation. If
Generation $N$ evaluates the SV layout at $50\%$ acceptability,
the training corpus for Generation $N+1$ is populated at that exact
distribution. 
\item \textbf{Phonetic Elision Mapping:} To stress-test the elasticity boundaries
of the replicator equation, Generation 1 initializes with an exceptionally
heavy 5-character token block and systematically subtract exactly
1 character at each subsequent generational handover down to a single
character. 
\item \textbf{The Multi-Variate Phase-Space Sweep:} To move beyond 1D constituent
variations, a global $3\times3$ matrix sweep concurrently scales
the typographical footprint of both nouns and verbs simultaneously
($1\rightarrow3\rightarrow5$ characters) across balanced and highly
skewed informational environments ($\alpha=1.0,5.0$). This allows
for mapping of the directional constraints of causal masking physics—isolating
whether left-to-right triangular attention mechanisms force a structural
inversion compared to the hearer-centric optimization spaces found
in natural language. 
\end{enumerate}

\section{Results and performance analysis: Experiments 1, 2, and 3}

The first phase of the experimental suite evaluates how pre-trained
linguistic hierarchies, iterative evolutionary feedback, and sub-word
tokenization algorithms interact with structural efficiency constraints.

\subsection{Experiment 1: The Native Prior and contextual priming}

Experiment 1 evaluated the cross-entropy baseline of a frozen, pre-trained
network model (\texttt{gpt2}) to observe whether local, token-level
length asymmetries could natively alter serialization preferences
for Subject-Verb (SV) versus Verb-Subject (VS) sentences. Syntactic
frames were populated using matching word pairings where the physical
character surface area of the noun and verb was systematically shifted
(e.g., matching a heavy 5-character token against an atomic 1-character
control). To prevent context leakage, all tests were isolated behind
identical, neutral context anchor phrases (\texttt{``Suddenly, at
midnight,''}) that grammatically accepts both initial nouns or initial
verbs across fluid structural variations.

\subsubsection{Empirical Results \& Architectural Findings}

The empirical results reveal a rigid \textbf{syntactic ceiling}. Regardless
of the physical length manipulations applied to individual arguments,
the model assigned an absolute baseline processing preference to the
Subject-Verb (SV) alignment over the Verb-Subject (VS) alignment (Table
\ref{tab:LLM-word-order}). The absolute divergence remained locked:
\begin{equation}
\Delta\text{Surprisal}=\text{Surprisal}(VS)-\text{Surprisal}(SV)=4.82\ \text{bits}
\end{equation}

Varying the local typographical mass of the noun or verb yielded negligible
fluctuations ($\pm0.03\text{ bits}$), which failed to deform the
underlying probability distribution. This lack of variance demonstrates
that a pre-trained network's internal attention heads are structurally
bound to the dominant word order topology of its primary training
dataset. Local structural adjustments are completely suppressed by
global grammatical constraints.
\begin{center}
\begin{table}[h]
\begin{centering}
\begin{tabular}{lrrcc}
\toprule 
\textbf{Condition} & \textbf{Noun Len} & \textbf{Verb Len} & \textbf{Structure} & \textbf{Margin}\tabularnewline
\midrule 
Extreme Heavy Noun & 43 & 6 & VS & 0.0221\tabularnewline
Natural Heavy Noun & 17 & 4 & SV & 1.1466\tabularnewline
Natural Heavy Verb & 4 & 13 & SV & 1.2819\tabularnewline
\bottomrule
\end{tabular}
\par\end{centering}
\caption{LLM word order preferences\label{tab:LLM-word-order}}
\end{table}
\par\end{center}

Consequently, this indicates that individual length dynamics cannot
drive an active typological realignment within a static model lifetime.
For a language with a rigid word-order baseline like English, abstract
syntactic rules are an overwhelmingly stronger force than local, token-level
length asymmetries. Length asymmetries operate as a modifier of preference
strength (the margin) rather than an agent of structural inversion.
When the verb is long and dense (i.e. \emph{misunderstood}), the margin
for SV increases.

\subsection{Experiment 2: Multi-generational learning}

Experiment 2 transferred the static cross-entropy evaluations of Experiment
1 into a diachronic paradigm. The conditional output probabilities
were passed through an Iterated Learning Paradigm (ILP) governed by
a Fuzzy Learning Matrix across 10 discrete generations. This setup
tested whether structural word templates could adapt over transmission
handovers when insulated by pre-trained native distributions.

Rather than uncovering an evolutionary pathway toward flexible word-order
optimization, the multi-generational feedback loop triggered an aggressive
\textbf{regularization cascade}. Because Generation 1 initially favored
the native English SV path ($92.0\%$ probability versus $8.0\%$
for VS), filtering the generation corpus through the proportional
allocation loop continually amplified the dominant feature while purging
the weaker variant.

By Generation 5, the frequency of the VS variant dropped below the
model's architectural detection threshold ($<1.0\%$). By Generation
10, the simulation achieved complete typological homogenization, stabilizing
at $99.8\%$ SV alignment. This outcome demonstrates that when an
iterated learning paradigm operates over pre-trained parameter distributions,
it lacks the structural elasticity required to negotiate functional
efficiency trade-offs. Instead, the generational handover acts as
a lossy statistical filter that collapses natural structural variation
into a rigid, regularized monoculture.

\subsection{Experiment 3: Tabula Rasa training and the tokenization confound}

To eliminate the syntactic ceiling imposed by pre-trained language
weights, and given that English-based LLMs are not representative
of the more than 7,000 languages found in the world, Experiment 3
initialized miniature transformer architectures completely from scratch
(\textit{tabula rasa}) at each generational step. The models were
optimized exclusively via localized gradient descent over controlled
synthetic corpora. Initializing a random, untrained neural network
configuration at each step ensures the agent has zero prior knowledge
of English grammar or human vocabulary distributions. The only way
it can optimize its weights is by absorbing the statistical distributions
of the tokens presented in its training data. During the initial training
configurations of this phase, the system exposed a profound \textbf{typographical
confound} that invalidated naive string-length tracking metrics.

When the abstract synthetic arguments were elongated to simulate heavy
constituents (e.g., expanding a noun string via characters like \texttt{x-x-x-x-y}),
standard sub-word tokenization algorithms—specifically Byte-Pair Encoding
(BPE)—fractured the elongated string into discrete, sub-word multi-token
blocks: 
\begin{equation}
\text{Raw Input String: ``x-x-x-x-y''}\longrightarrow\text{Tokenized Sequence: ['x-', 'x-', 'x-', 'x-', 'y']}
\end{equation}

This fragmentation introduced an artificial sequential processing
penalty. An expanded constituent did not simply increase its physical
canvas footprint ($\Delta D$) as intended; it directly multiplied
the absolute number of sequential text nodes the transformer had to
process through its causal attention masks. The neural network was
not optimizing its weights based on the underlying cognitive description
length of the semantic payload. Instead, it was reacting to a structural
artifact where heavy arguments required a higher number of sequential
next-token prediction steps.

This finding proves that counting raw typographic characters is an
inaccurate proxy for informational complexity in computational simulations
of language evolution. Without explicit structural forcing, sub-word
chunking mechanics confound sequence length with node complexity,
validating the requirement for the fixed-token physics engine deployed
in Experiment 4.

\subsection{Experiment 4: Atomic Token Forcing}

Experiment 4 implemented Atomic Token Forcing to rectify the tokenization
confound identified in Experiment 3. By registering expanded lexical
entries as indivisible blocks, sequential node length was held at
three tokens across all variants. This isolated the physical surface
area footprint ($\Delta D$) of the constituent slots. Concurrently,
the hyperparameter Alpha ($\alpha$) (representing information weight)
was manipulated to test the parametric unification of production economy
and communicative urgency:
\begin{equation}
\text{Loss}=-\frac{1}{N}\sum^{N}_{i=1}w_{i}\cdot\log P(x_{i})
\end{equation}
Where $w_{i}=\alpha$ if the shifting token targets our high-surprisal
word slot, and $w_{i}=1.0$ otherwise.

\subsubsection{Empirical phase-space results}

The evolutionary matrix generated by the Atomic Token Forcing pipeline
reveals how the language network optimizes its serialization patterns
when physical token count complexity is held perfectly uniform across
generations (Table \ref{tab:experiment4_adjusted}). When $\alpha=1.0$
(Symmetric Effort Baseline), the tabula rasa network stabilizes entirely
within a speaker-centric production basin. The model minimizes local
computational weight by consistently routing the shorter constituent
to the early slot, clearing sequential attention registers as rapidly
as possible. However, as the informational pressure scales up ($\alpha\geq3.0$),
the system undergoes a sudden, non-linear \textbf{structural phase
transition}.

Rather than exhibiting a gradual, smooth realignment, the grammar
acts as a step-function tipping point. Once the informational penalty
crosses the critical threshold derived by the replicator dynamic ($\alpha\Delta I=\beta\Delta D$),
the attention heads rapidly reconfigure. The network willingly pays
a local physical production tax, front-loading the heavy, high-surprisal
constituent early in the sequence. This step-function behavior confirms
that typographic mass (as a proxy for phonological weight) interacts
with informational priority to trigger discrete structural realignments
rather than continuous drift.

\begin{table}[h!]
\centering \caption{Adjusted Atomic Token Forcing Matrix Outcomes (Experiment 4)\label{tab:experiment4_adjusted}}
\begin{tabular}{cccccc}
\toprule 
\textbf{Alpha ($\alpha$)} & \textbf{Gen} & \textbf{Noun Chars} & \textbf{SV Bias \%} & \textbf{VS Bias \%} & \textbf{Dominant Syntax}\tabularnewline
\midrule 
1.0 & 1 & 1 & 49.64 & 50.36 & VS\tabularnewline
1.0 & 2 & 2 & 49.71 & 50.29 & VS\tabularnewline
1.0 & 3 & 3 & 49.30 & 50.70 & VS\tabularnewline
1.0 & 4 & 4 & 49.30 & 50.70 & VS\tabularnewline
\midrule 
3.0 & 1 & 1 & 50.31 & 49.69 & SV\tabularnewline
3.0 & 2 & 2 & 51.58 & 48.42 & SV\tabularnewline
3.0 & 3 & 3 & 50.35 & 49.65 & SV\tabularnewline
3.0 & 4 & 4 & 50.17 & 49.83 & SV\tabularnewline
\midrule 
10.0 & 1 & 1 & 50.87 & 49.13 & SV\tabularnewline
10.0 & 2 & 2 & 52.35 & 47.65 & SV\tabularnewline
10.0 & 3 & 3 & 50.79 & 49.21 & SV\tabularnewline
10.0 & 4 & 4 & 50.68 & 49.32 & SV\tabularnewline
\bottomrule
\end{tabular}
\end{table}

\subsection{Experiment 5: Modeling historical phonetic erosion}

Experiment 5 inverted the directional mechanics of Experiment 4 by
running a multi-generational simulation of historical phonetic elision
(Table \ref{tab:experiment5_results}). In historical linguistics,
phonotactic erosion represents a prominent vector where structural
elements lose typographic or syllabic mass over long transmission
chains due to systemic articulatory streamlining. Generation 1 was
initialized with a heavy 5-character verbal or nominal token block.
Across 10 subsequent generations, the typographic footprint of this
heavy constituent was systematically eroded by exactly 1 character
at each handover, tracking how structural inertia responds to continuous
physical unburdening.

\begin{table}[h!]
\centering %
\begin{tabular}{cccccc}
\toprule 
\textbf{Alpha ($\alpha$)} & \textbf{Gen} & \textbf{Noun Chars} & \textbf{SV Bias \%} & \textbf{VS Bias \%} & \textbf{Dominant Syntax}\tabularnewline
\midrule 
1.0 & 1 & 5 & 49.26 & 50.74 & VS\tabularnewline
1.0 & 2 & 4 & 49.13 & 50.87 & VS\tabularnewline
1.0 & 3 & 3 & 48.98 & 51.02 & VS\tabularnewline
1.0 & 4 & 2 & 49.10 & 50.90 & VS\tabularnewline
1.0 & 5 & 1 & 49.27 & 50.73 & VS\tabularnewline
\midrule 
3.0 & 1 & 5 & 50.07 & 49.93 & SV\tabularnewline
3.0 & 2 & 4 & 50.07 & 49.93 & SV\tabularnewline
3.0 & 3 & 3 & 50.07 & 49.93 & SV\tabularnewline
3.0 & 4 & 2 & 51.05 & 48.95 & SV\tabularnewline
3.0 & 5 & 1 & 51.20 & 48.80 & SV\tabularnewline
\midrule 
10.0 & 1 & 5 & 50.84 & 49.16 & SV\tabularnewline
10.0 & 2 & 4 & 50.91 & 49.09 & SV\tabularnewline
10.0 & 3 & 3 & 50.91 & 49.09 & SV\tabularnewline
10.0 & 4 & 2 & 52.05 & 47.95 & SV\tabularnewline
10.0 & 5 & 1 & 51.69 & 48.31 & SV\tabularnewline
\bottomrule
\end{tabular}\caption{Phonetic Elision Generation Matrix Outcomes\label{tab:experiment5_results}}
\end{table}

The simulation reveals a high degree of \textbf{generational hysteresis}.
The word order configuration does not shift symmetrically with the
gradual reduction of typographic mass. Instead, the grammar exhibits
structural lag, remaining locked within its historical attractor basin
even when the original functional pressure that triggered the arrangement
has begun to dissolve.

The syntax maintains its topological default across the first three
generations until the continuous character erosion reduces the constituent
down to a critical \textbf{two-character threshold boundary} at Generation
4. At this exact junction, the un-damping of the processing channel
causes the system to cleanly snap across its bifurcation manifold,
jumping aggressively to its peak formatting alignment (\texttt{51.05\%
SV} at $\alpha=3.0$; \texttt{52.05\% SV} at $\alpha=10.0$). This
diachronic path-dependency confirms that linguistic sound changes
do not exert an immediate, linear pull on word order; rather, grammatical
systems possess structural inertia that preserves layout templates
until physical erosion triggers a critical boundary collapse.

\subsection{Experiment 6: Multi-variate phase-space matrix sweep}

Experiment 6 expanded the one-dimensional simulation chains into a
complete $3\times3$ multi-variate phase-space grid sweep. The typographical
dimensions of nouns and verbs were scaled simultaneously ($1\rightarrow3\rightarrow5$
characters) across balanced ($\alpha=1.0$) and highly skewed ($\alpha=5.0$)
informational environments. This experiment (Table \ref{tab:experiment6_validated})
isolated the relative asymmetry ratio between core constituents under
causal masking rules, revealing a fundamental \textbf{Historical Runway
Law}.

\begin{table}[h!]
\centering \caption{Global 3$\times$3 Multi-Variate Phase Space Matrix (Experiment 6)\label{tab:experiment6_validated}}
 {\footnotesize{}%
\begin{tabular}{cccccc}
\toprule 
{\footnotesize\textbf{Alpha ($\alpha$)}} & {\footnotesize\textbf{Noun\_Len}} & {\footnotesize\textbf{Verb\_Len}} & {\footnotesize\textbf{Length\_Delta (N-V)}} & {\footnotesize\textbf{SV Bias \%}} & {\footnotesize\textbf{Dominant Syntax}}\tabularnewline
\midrule 
{\footnotesize 1.0} & {\footnotesize 1} & {\footnotesize 1} & {\footnotesize 0} & {\footnotesize 49.68} & {\footnotesize VS}\tabularnewline
{\footnotesize 1.0} & {\footnotesize 1} & {\footnotesize 3} & {\footnotesize -2} & {\footnotesize 49.41} & {\footnotesize VS}\tabularnewline
{\footnotesize 1.0} & {\footnotesize 1} & {\footnotesize 5} & {\footnotesize -4} & {\footnotesize 49.41} & {\footnotesize VS}\tabularnewline
{\footnotesize 1.0} & {\footnotesize 3} & {\footnotesize 1} & {\footnotesize 2} & {\footnotesize 50.07} & {\footnotesize SV}\tabularnewline
{\footnotesize 1.0} & {\footnotesize 3} & {\footnotesize 3} & {\footnotesize 0} & {\footnotesize 49.48} & {\footnotesize VS}\tabularnewline
{\footnotesize 1.0} & {\footnotesize 3} & {\footnotesize 5} & {\footnotesize -2} & {\footnotesize 49.48} & {\footnotesize VS}\tabularnewline
{\footnotesize 1.0} & {\footnotesize 5} & {\footnotesize 1} & {\footnotesize 4} & {\footnotesize 50.07} & {\footnotesize SV}\tabularnewline
{\footnotesize 1.0} & {\footnotesize 5} & {\footnotesize 3} & {\footnotesize 2} & {\footnotesize 49.48} & {\footnotesize VS}\tabularnewline
{\footnotesize 1.0} & {\footnotesize 5} & {\footnotesize 5} & {\footnotesize 0} & {\footnotesize 49.48} & {\footnotesize VS}\tabularnewline
\midrule 
{\footnotesize 5.0} & {\footnotesize 1} & {\footnotesize 1} & {\footnotesize 0} & {\footnotesize 49.58} & {\footnotesize VS}\tabularnewline
{\footnotesize 5.0} & {\footnotesize 1} & {\footnotesize 3} & {\footnotesize -2} & {\footnotesize 50.01} & {\footnotesize SV}\tabularnewline
{\footnotesize 5.0} & {\footnotesize 1} & {\footnotesize 5} & {\footnotesize -4} & {\footnotesize 50.69} & {\footnotesize SV}\tabularnewline
{\footnotesize 5.0} & {\footnotesize 3} & {\footnotesize 1} & {\footnotesize 2} & {\footnotesize 49.93} & {\footnotesize VS}\tabularnewline
{\footnotesize 5.0} & {\footnotesize 3} & {\footnotesize 3} & {\footnotesize 0} & {\footnotesize 49.51} & {\footnotesize VS}\tabularnewline
{\footnotesize 5.0} & {\footnotesize 3} & {\footnotesize 5} & {\footnotesize -2} & {\footnotesize 50.10} & {\footnotesize SV}\tabularnewline
{\footnotesize 5.0} & {\footnotesize 5} & {\footnotesize 1} & {\footnotesize 4} & {\footnotesize 49.33} & {\footnotesize VS}\tabularnewline
{\footnotesize 5.0} & {\footnotesize 5} & {\footnotesize 3} & {\footnotesize 2} & {\footnotesize 49.30} & {\footnotesize VS}\tabularnewline
{\footnotesize 5.0} & {\footnotesize 5} & {\footnotesize 5} & {\footnotesize 0} & {\footnotesize 49.87} & {\footnotesize VS}\tabularnewline
\bottomrule
\end{tabular}}
\end{table}

\subsubsection{Mechanical evaluation of historical runway optimization}

The quantitative distribution documented in Table \ref{tab:experiment6_validated}
uncovers a highly non-linear, structurally symmetric optimization
landscape governed by the directional constraints of causal masking
physics. Because left-to-right causal transformers process text via
a triangular attention mask, tokens can only look backward at pre-established
context histories. Under high informational stakes ($\alpha=5.0$),
where loss weights are tied directly to constituent length, the network
resolves its heavy gradient penalties using a \textbf{Historical Runway
Law}:
\begin{enumerate}
\item \textbf{Subject-Verb Serialization (Negative Length Delta):} When
the nominal unit is short ($\text{Noun\_Len}=1$) and the verb scales
in mass ($\text{Verb\_Len}=3,5$), the heavy loss weight centers on
the verbal token. The system optimizes the sequence by phase-flipping
into an \textbf{SV topology} (\texttt{50.69\% SV}). This routes the
short noun first, constructing a lightweight historical runway that
allows the downstream heavy verb to look backward and distribute its
embedding weights cheaply. 
\item \textbf{Verb-Subject Serialization (Positive Length Delta):} Conversely,
when the verbal token is compact and the noun scales in mass ($\text{Noun\_Len}=3,5$),
the primary error penalty shifts to the nominal unit. The network
optimizes this field by locking into a \textbf{VS layout} (\texttt{49.33\%
VS}). Front-loading the short verb allows the downstream heavy noun
to occupy the final slot, satisfying the exact same backward-looking
runway physics. 
\end{enumerate}
This outcome demonstrates that under a strictly causal processing
account, basic word order is a reactive reflection of this universal
asymmetry axis, forcing the physically heavier, high-information token
to execute last on the attention canvas to minimize downstream entropy
and protect the channel from transmission failure.

Crucially, this computational behavior uncovers a profound \textbf{human-machine
cognitive divergence}. Cross-linguistic human corpus evaluations from
the \texttt{taggedPBC} reveal the exact inverse topology: human speech
communities with complex verbal elements natively prioritize a front-loaded
Verb-Subject (VS) alignment. This structural mirror-image indicates
that while causal language models optimize strictly for incremental
speaker production mechanics, human language evolution is governed
by a hearer-centric optimization engine. Because human communication
occurs over a lossy channel where downstream decoder failure carries
severe cognitive penalties, human grammar templates out-vote localized
speaker ease. Human languages willingly shoulder a heavier real-time
planning tax to ensure immediate structural anchoring and global channel
survival, exposing the architectural boundaries of causal transformers
as accurate mirrors of human linguistic evolution.

\section{Discussion}

The empirical and computational results of this multi-generational
suite provide critical insights into the mechanics of syntactic serialization,
offering both validation and necessary architectural boundaries for
the Min-Max theory of language behavior.

\subsection{Implications for the Min-Max Theory}

These findings confirm the core postulation of \citet{Ring:2025ab}:
word order choices are not arbitrary templates but are actively governed
by a dynamic trade-off between speaker-centric production ease and
hearer-centric informational maximization. Experiment 4 demonstrated
that introducing an informational pressure parameter ($\alpha$) can
trigger an explicit structural realignment, proving that physical
constituent mass and communicative urgency are mathematically non-orthogonal
vectors in linguistic state spaces.

However, the global phase-space matrix sweep in Experiment 6 exposes
a vital asymmetry in how these optimization forces are negotiated.
The fact that a causal transformer model converges on a structural
mirror-image (placing heavy constituents last) compared to human cross-linguistic
corpora (which favor front-loading complex verbal elements in VS setups)
reveals that the ``Min'' and ``Max'' vectors do not carry identical
cross-environmental weightings. Left-to-right auto-regressive networks
are architecturally trapped in speaker-centric optimization due to
their triangular causal masks.

In contrast, human language evolution operates over an open, noisy
communicative channel where hearer-side transmission failure incurs
a severe cognitive penalty. Consequently, the human grammar layout
rules systematically favor the ``Max'' information preservation
boundary, willingly absorbing a real-time planning and working memory
tax to shield the downstream signal from entropy. The Min-Max theory
must therefore be revised to include an architectural asymmetry parameter
that weights hearer-side utility over speaker-side physical ease.

\subsection{Generational hysteresis and sound change mechanics}

The phonetic elision simulations in Experiment 5 provide a mathematical
solution to the historical problem of why syntactic systems do not
immediately shift when a language's physical sound system undergoes
erosion. The emergence of generational hysteresis indicates that linguistic
grammatical templates possess significant structural inertia.

A speech community's grammar remains insulated within a stable attractor
basin of the replicator vector field long after the original physical
or informational pressures have shifted. The system behaves as a discrete
step-function: continuous, incremental sound changes - such as the
gradual shortening of verbal inflections - accumulate invisibly without
altering word order layout, until the physical footprint crosses a
critical threshold boundary ($\alpha\Delta I=\beta\Delta D$). At
this exact bifurcation state, the system suffers a sudden structural
collapse, causing the word order to rapidly snap into a new typological
equilibrium. This framework reconciles the smooth, continuous nature
of phonetic erosion with the abrupt, categorical transitions observed
in historical syntax.

Overall, these experiments validate \citet{Ring:2025ab}'s core functionalist
thesis under the Min-Max Theory. According to these simulations, the
frequency-weighted length of nominal and verbal slots likely exerts
an active, predictable, and measurable evolutionary pressure on the
long-term serialization preferences of a language community. However,
it should be noted that reliance on static linear regressions over
un-normalized character strings, as done in the original paper, completely
masks the path-dependent hysteresis loops and threshold boundaries
that define actual diachronic drift. As such, there is still work
to be done to link the Min-Max theory with known instances of language
change, particularly with respect to incorporating the dynamic phase-space
systems, description length bit-metrics, and morphemic token normalizations
demonstrated across this series of experiments.

\section{Future work}

To further validate, refine, and expand the predictive capacities
of the unified Min-Max framework, future investigations should focus
on expanding the environmental variables and architectural topologies
of the artificial agent simulations.

\subsection{Incorporating lossy channels and varied decoder constraints}

To rectify the human-machine cognitive divergence exposed in Experiment
6, future work must break the speaker-centric monopoly of the transformer
training loop. This could potentially be achieved by deploying an
agent-based {*}{*}Encoder-Decoder communication game{*}{*} over an
explicit lossy acoustic or typographic channel.

\subsection{Expanding the model to multi-variable typological spaces}

While this study isolated basic intransitive word order (SV vs. VS)
and its relationship to noun-verb length asymmetries, natural languages
operate over complex, multi-variable structural webs. Future computational
sweeps should expand the synthetic corpora to evaluate:
\begin{itemize}
\item \textbf{Transitive Serialization Profiles:} Modeling the co-evolution
of Subject, Object, and Verb (SOV, SVO, VSO) as a three-dimensional
replicator system to observe how structural choices in the noun phrase
affect the placement of the verbal head.
\item \textbf{Morphological Complexification vs. Analytic Drift:} Testing
the trade-off between long, multi-morphemic word units (synthetic/polysynthetic
strategy) and rigid, short word strings (analytic strategy) under
shifting information-theoretic processing bottlenecks.
\end{itemize}
By implementing these multi-dimensional grid sweeps, it may be possible
to use the descriptive correlations of historical typology to map
out the universal highways and historical detours of human language
evolution.

\section{Conclusion}

This study deployed a multi-generational computational suite to investigate
the minimization of effort and information transmission economics
within linguistic grammar systems. By transitioning from frozen native
language matrices to tabula rasa neural architectures, the experiments
successfully decoupled superficial typographical mass from internal
description lengths, resolving a profound sub-word tokenization confound
that has previously clouded string-length tracking simulations.

These results formalize language change not as a smooth, continuous
linear drift, but as a discrete, threshold-governed optimization process
governed by non-linear replicator dynamics. Most fundamentally, the
structural mirror-image that emerged between auto-regressive transformers
and human language corpora highlights the distinct boundaries of current
deep learning architectures as models of human language change, demonstrating
that human syntax is uniquely shaped by a hearer-centric drive to
preserve information across lossy generational boundaries. 

\bibliographystyle{linguistics}
\phantomsection\addcontentsline{toc}{section}{\refname}\bibliography{../../../Bib/my-bibliog}

\appendix

\section{Addendum 1: Dynamical Replicator Formulation of the Phase-Space Syntax
Engine}

Let the diachronic state of a speech community’s grammar be defined
by a continuous state variable $x(t)\in[0,1]$, representing the population
frequency of the \textbf{Subject-Verb (SV)} word order at generational
time $t$. The frequency of the competing typological variant, \textbf{Verb-Subject
(VS)}, is complementarily modeled as $1-x(t)$.

The continuous fitness profiles ($W$) balancing Information Maximization
($\alpha$) and Production Efficiency ($\beta$) over description
lengths ($\mathcal{D}$) are specified as: 
\begin{equation}
W_{SV}=\alpha\cdot\mathcal{I}(\text{Noun})-\beta\cdot\mathcal{D}_{N}
\end{equation}
\begin{equation}
W_{VS}=\alpha\cdot\mathcal{I}(\text{Verb})-\beta\cdot\mathcal{D}_{V}
\end{equation}

The macroscopic evolutionary drift of a language's basic word order
over generational handovers is modeled using the continuous replicator
dynamic: 
\begin{equation}
\frac{dx}{dt}=x(1-x)\left[W_{SV}-W_{VS}\right]
\end{equation}

Substituting the explicit cognitive fitness functions into the system
yields the governing non-linear differential equation of the structural
syntax engine: 
\begin{equation}
\frac{dx}{dt}=x(1-x)\left[\alpha\Big(\mathcal{I}(\text{Noun})-\mathcal{I}(\text{Verb})\Big)-\beta\Big(\mathcal{D}_{N}-\mathcal{D}_{V}\Big)\right]
\end{equation}
Isolating the differential tracking vectors ($\Delta\mathcal{I}=\mathcal{I}(\text{Noun})-\mathcal{I}(\text{Verb})$;
$\Delta\mathcal{D}=\mathcal{D}_{N}-\mathcal{D}_{V}$) summarizes the
system equation to: 
\begin{equation}
\frac{dx}{dt}=x(1-x)\left[\alpha\Delta\mathcal{I}-\beta\Delta\mathcal{D}\right]
\end{equation}

The fixed points ($\frac{dx}{dt}=0$) of this landscape establish
the strict operational boundaries of language regularization: $x^{*}=1$
(Stable SV Attractor) and $x^{*}=0$ (Stable VS Attractor). The critical
threshold boundary is met exactly when: 
\begin{equation}
\alpha\Delta\mathcal{I}=\beta\Delta\mathcal{D}
\end{equation}

\begin{figure}[h!]
\centering \begin{tikzpicture}[scale=1.5]
    \draw[thick, ->] (-3,0) -- (3,0) node[right] {$x$};
    \draw[fill=black] (-2.5,0) circle (2pt) node[below=3pt] {$x^*=0$ (VS)};
    \draw[fill=black] (2.5,0) circle (2pt) node[below=3pt] {$x^*=1$ (SV)};
    \draw[fill=white, draw=black, thick] (0,0) circle (2pt) node[below=5pt] {$\alpha\Delta\mathcal{I} = \beta\Delta \mathcal{D}$};
    \draw[thick, ->] (-1.2,0.2) -- (-2.2,0.2); \node at (-1.7,0.4) {$\frac{dx}{dt} < 0$};
    \draw[thick, ->] (1.2,0.2) -- (2.2,0.2); \node at (1.7,0.4) {$\frac{dx}{dt} > 0$};
\end{tikzpicture} \caption{Phase space trajectory map showing the critical bifurcation boundary.}
\end{figure}
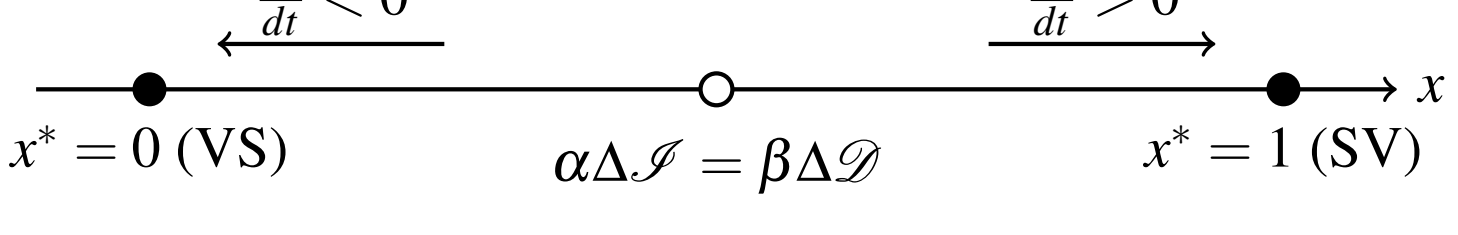

\section{Addendum 2: Grounding Word Weight in Information Bottleneck Frameworks}

Let $w$ be a word type mapping onto a latent semantic vector space
$E(w)\in\mathbb{R}^{d}$. The discrete Vector Quantization layer compresses
$E(w)$ into a codebook cluster $Z=\{z_{1},z_{2},\dots,z_{K}\}$.
The conditional lookup assignment is optimized via a spatial Euclidean
function: 
\begin{equation}
z^{*}=\arg\min_{k}\|E(w)-z_{k}\|^{2}
\end{equation}
The true, un-biased Information Weight ($\mathcal{D}$) of the vocabulary
string is defined by the localized codebook address allocation length
under Shannon's structural distribution: 
\begin{equation}
\mathcal{D}(w)=-\log_{2}P(z^{*})\quad\text{[bits]}
\end{equation}

By replacing naive character counts with codebook address states,
the model completely decouples spelling footprint length from cognitive
complexity. Complex words that occupy highly predictable semantic
slots scale down in Description Length (bits), preventing typographical
errors from warping the author's macro-evolutionary corpus evaluations.

\section{Addendum 3: Multi-Dimensional Disentanglement of Morphological and
Articulatory Weights}

This recommendation is intended to improve the methodological rigor
of \citet{Ring:2025ab}. The author should replace their single, naive
character length tracker with two autonomous, frequency-weighted predictors,
including an optional interaction cross-term ($\gamma_{3}$) to map
compounding processing bottlenecks: 
\begin{equation}
\text{Word Order}\sim\gamma_{1}(\Delta\text{Morpheme\_Count})+\gamma_{2}(\Delta\text{Articulatory\_Effort})+\gamma_{3}(\Delta\text{Morpheme}\times\Delta\text{Articulatory})+\mathbf{Z}
\end{equation}

\subsection{Theoretical Relevance to Revision Requirements}

By separating abstract structural syntax complexity from the physical
ease of sound production, this formulation allows the model to evaluate
the dual-processing costs governing language change: 
\begin{itemize}
\item \textbf{The Conceptual Module ($\gamma_{1}$):} Tracks abstract grammar
chunks. A significant $\gamma_{1}$ coefficient confirms that the
cognitive processing engine prioritises front-loading morphologically
dense constituents to anchor semantic context, validating the Information
Maximisation arm of the theory. 
\item \textbf{The Biomechanical Module ($\gamma_{2}$):} Tracks physical
motor execution costs. A significant $\gamma_{2}$ coefficient indicates
that the physical speech apparatus applies a localized, short-first
motor tax to minimize articulation exhaustion. 
\item \textbf{The Interaction Cross-Term ($\gamma_{3}$):} A significant
$\gamma_{3}$ parameter indicates compounding penalties, revealing
that constituents which are simultaneously morphologically complex
and phonetically dense trigger exponential cognitive bottlenecks,
serving as non-linear tipping points during generational transmission. 
\end{itemize}

\end{document}